\documentclass[letterpaper, 10pt, conference]{ieeeconf}

\usepackage{times}
\usepackage{tabularx}
\usepackage{subcaption}
\usepackage{multirow,multicol, array}
\usepackage[font={small}]{caption}   
\usepackage{graphicx,dblfloatfix}
\usepackage{wrapfig}
\usepackage{siunitx}

\let\labelindent\relax
\usepackage{enumitem}

\usepackage{amsmath,amsthm,amssymb,amsfonts}
\usepackage[titlenumbered,ruled]{algorithm2e}
\usepackage{textcomp}
\usepackage{acronym}
\usepackage{balance}
\usepackage{mdwmath}
\usepackage{bm}
\usepackage{mdwtab}
\usepackage{array}
\usepackage[usenames,dvipsnames]{color}
\usepackage{eqparbox}
\usepackage{cite}

\usepackage{color}
\usepackage{psfrag}
\usepackage{epsfig}
\usepackage{url}

\usepackage{epstopdf}
\usepackage{booktabs}
\usepackage{blindtext}

\usepackage[colorinlistoftodos,prependcaption,textsize=tiny]{todonotes}

\renewcommand{\emph}{\textit}

\newtheorem*{lemma*}{Lemma}

\newtheorem*{problem*}{Problem}

\makeatletter
\newcommand\fs@spaceruled{\def\@fs@cfont{\bfseries}\let\@fs@capt\floatc@ruled
    \def\@fs@pre{\vspace{5\baselineskip}\hrule height.8pt depth0pt \kern2pt}%
    \def\@fs@post{\kern2pt\hrule\relax}%
    \def\@fs@mid{\kern2pt\hrule\kern2pt}%
    \let\@fs@iftopcapt\iftrue}
\makeatother

\IEEEoverridecommandlockouts
\graphicspath{{images/}}

\begin{document}
	\acrodef{ISR}[\textsc{isr}]{Intelligence, Surveillance, and Reconnaissance}
	
	\title{Motion Planning for Collision-resilient Mobile Robots in Obstacle-cluttered Unknown Environments with Risk Reward Trade-offs}
	
	\author{Zhouyu Lu, Zhichao Liu, Gustavo J. Correa, and Konstantinos Karydis
		\thanks{The authors are with the Dept. of Electrical and Computer Engineering, University of California, Riverside. 
			Email: {\{zlu044, zliu157, gcorr003, karydis\}@ucr.edu}. 
		We gratefully acknowledge the support of NSF under grant \#IIS-1910087, ONR under grants \#N00014-18-1-2252 and \#N00014-19-1-2264, ARL under grant
		\#W911NF-18-1-0266, and UCR Office of Research and Economic Development under a Collaborative Seed Grant. Any opinions, findings, and conclusions or recommendations expressed in this material are those of the authors and do not necessarily reflect the views of the funding agencies.}
	}

	\maketitle
	\thispagestyle{empty}

	\begin{abstract}
	    Collision avoidance in unknown obstacle-cluttered environments may not always be feasible.  This paper focuses on an emerging paradigm shift in which potential collisions with the environment can be harnessed instead of being avoided altogether. To this end, we introduce a new sampling-based online planning algorithm that can explicitly handle the risk of colliding with the environment and can switch between collision avoidance and collision exploitation. Central to the planner's capabilities is a novel joint optimization function that evaluates the effect of possible collisions using a reflection model. This way, the planner can make deliberate decisions to collide with the environment if such collision is expected to help the robot make progress toward its goal. To make the algorithm online, we present a state expansion pruning technique that significantly reduces the search space while ensuring completeness. The proposed algorithm is evaluated experimentally with a built-in-house holonomic wheeled robot that can withstand collisions. We perform an extensive parametric study to investigate trade-offs between (user-tuned) levels of risk, deliberate collision decision making, and trajectory statistics such as time to reach the goal and path length. 
	    %
	    %
	\end{abstract}

	\section{Introduction}
	Achieving high mission performance in face of risk is a common trade-off in robot motion planning~\cite{xiao2019explicit}. The trade-off becomes increasingly important for planning and navigation of robots with noisy actuation and sensing (such as small and inexpensive robots), especially when there is need to deploy them rapidly into dangerous or otherwise not easily accessible areas. A key source of risk includes possible collisions when navigating in an unknown environment. 
	
	Autonomous navigation in unknown environments can impact several applications ranging from surveying and
	inspection to search and rescue~\cite{tordesillas2019real}, and has thus received significant attention. Conventional planning and navigation strategies have used geometric representations of boundaries to define collision free paths with convergence guarantees~\cite{lavalle2006planning}. In partially-known environments, planning with local information based on instantaneous sensor data for collision avoidance is appealing for computationally-constrained platforms requiring high-rate collision avoidance~\cite{lopez2017aggressive}. Three main categories exist for local collision avoidance planning algorithms~\cite{oleynikova2018safe}, which include planning by reacting to sensor data~\cite{vasilopoulos2018reactive}, map-based planning using sensors to build maps or using a priori known global maps~\cite{pivtoraiko2013incremental, dhawale2018reactive}, and planning to maximize exploration coverage~\cite{heng2014autonomous, davis2016c}. 
	
	However, as robots increasingly venture outside the protected lab environment and into the real---uncertain---world, guaranteeing collision avoidance becomes an even more challenging task~\cite{hoy2015algorithms, campbell2012review}. If a conservative local planner is employed to avoid collisions in cluttered environments, the robot may not be able to find a feasible path to the goal even if one exists. At the same time, recent advances in material science and mechanical design have helped introduce robots that can safely withstand collisions (e.g., small legged robots with exoskeletons~\cite{haldane2015integrated}, aerial robots with protective cages~\cite{briod2014collision}, and soft robots~\cite{li2019agile}).

\begin{figure}[!t]
\vspace{6pt}
      \centering
      \begin{subfigure}{0.235\textwidth}
        \includegraphics[trim={3cm 0.2cm 3cm 0.2cm}, clip, width=\textwidth]{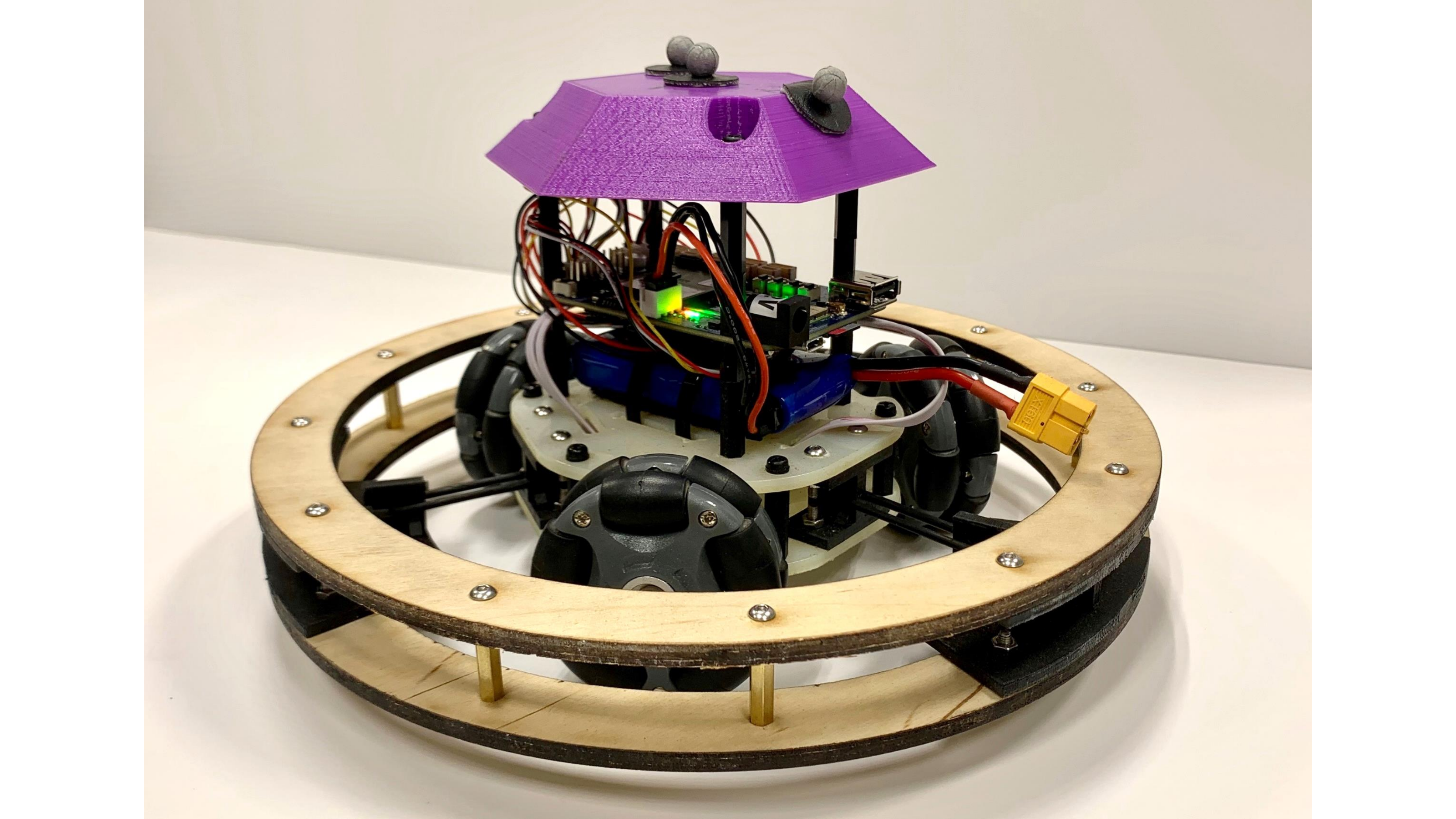}
      \end{subfigure}
      \begin{subfigure}{0.235\textwidth}
        \includegraphics[trim={3cm 0.2cm 3cm 0.2cm}, clip, width=\textwidth]{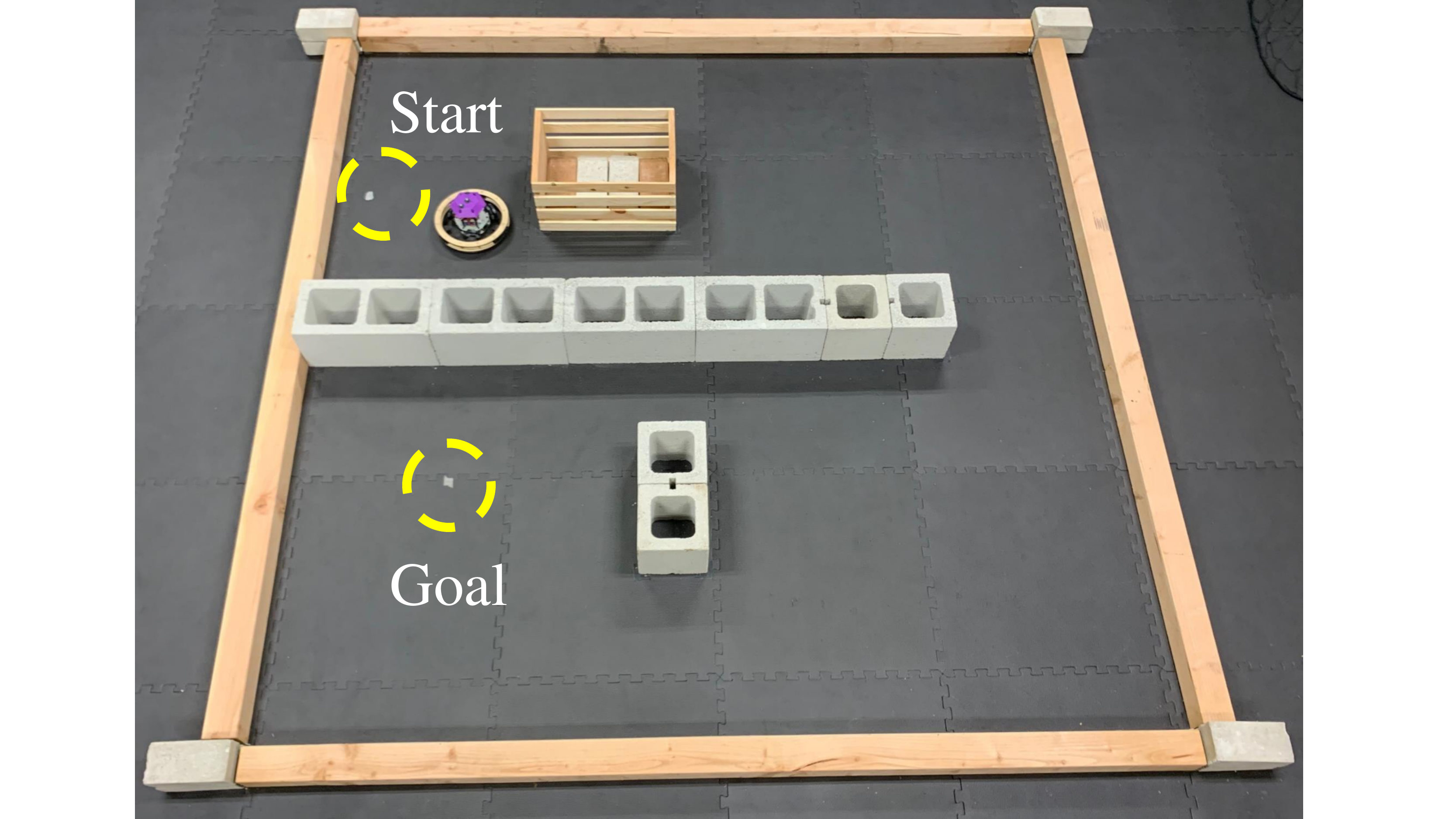}
      \end{subfigure}
      \caption{This study's holonomic wheeled robot (left) navigating in a confined corridor environment populated with obstacles (right) while \emph{harnessing} collisions. A supplementary video showing instances of the experiments can be found in  \protect\url{https://www.youtube.com/watch?v=S3oYebJRfA0}.}
      \label{fig:introFIG}
	\vspace{-16pt}
\end{figure}

Taken together, these observations may explain an emerging paradigm shift: \emph{Collisions with the environment could be harnessed instead of being avoided}. In fact, it has been demonstrated how allowing for collisions can benefit motion planning and control~\cite{karydis2014planning}, localization~\cite{mayya2018localization}, sensing~\cite{schmickl2009get}, and robot agility in terms of rapidly changing direction of motion~\cite{haldane2016robotic}. Collisions can also be useful for exploration of cluttered unknown environments~\cite{mulgaonkar2017robust}. 
Our work in this paper focuses on online motion planning that balances collision avoidance and harnessing collisions for mobile robot navigation in unknown environments.
	
Understanding the trade-offs and balancing between performance and risk has been a research focus for a while. One approach is to utilize the concept of inevitable collision states (ICS) for safe path planning~\cite{fraichard2004inevitable}, where the states that cannot avoid future collision must be prohibited. 
However, identifying appropriate ICS can be challenging for robots with noisy actuation and sensing.  
A different approach is to generate an explicit risk function based only on partial information of the robot state i.e. its location with respect to obstacles, while ignoring its speed and heading~\cite{pfeiffer2009probabilistic, liu2016episodic, huang2017viable}. However, ignoring the kinematic and dynamic properties of the robot may yield imprecise results. 
Merging these two types of approaches is possible through a continuous risk function 
based on the concept of collision time, which is the time in which the vehicle can hit the obstacle along its heading, if it were to lose control~\cite{song2018t}.
	
	While current risk-aware approaches can describe trade-offs between risk and reward of collision avoidance in mobile robot navigation, they cannot help determine \emph{how to best utilize those collisions in unknown environments}. 
	In related yet distinct efforts, models for reflection on obstacles have been developed to capture velocity, position, and uncertainties that result to collisions~\cite{StagerISER18,mulgaonkar2017robust,StagerICRA19}.  These works can be leveraged to explicitly include boundary interactions within planning algorithms. 
When the environment is known, collisions can serve as a practical means to improve the effectiveness of trajectories; through dissipation of energy or redirection of momentum, colliding agents can thus be endowed with greater maneuverability~\cite{mote2020collision}.  
Moreover, accepting collisions can help relax some of the safety conditions when generating a (conservative) path based on ICS planning; the latter may in cases make reaching a desired goal position slower~\cite{tordesillas2019real}. 
In previous work~\cite{lu2019optimal}, we showed how wall-following maneuvers can benefit a stochastically-moving robot to reach its goal in ingress/egress tasks. 
	
In this paper, we design a reactive sampling-based planner for collision-resilient mobile robots navigating in unknown obstacle-cluttered environments. The planner relies only on local information about the environment gathered through its sensors. We first propose a reward function to evaluate the effect of possible collisions based on a reflection model for the robot. We define a risk function based on the concept of collision time. The reward function is integrated with the risk function and a distance metric cost into a joint optimization function for motion planning. 
	
	The main contributions of the paper are as follows. 
	\begin{itemize}
		\item We present a planner capable of evaluating the trade-offs between harnessing and avoiding collisions in unknown environments. 
		\item We propose a novel formulation of reward function based on a reflection model to help the robot utilize collisions in motion planning.
		\item We present a pruning technique that can significantly reduce the search space while maintaining the solution quality and ensuring completeness. 
	\end{itemize}
	
	The proposed algorithm is tested with the Omnipuck holonomic wheeled robot~\cite{StagerISER18}, which we built in-house. The robot includes a reflection ring to help withstand collisions and rapidly change direction of motion after a collision by passively redirecting impact energy. 
	The algorithm is tunable by the user, and thus able to produce less risky paths at the expense of increasing overall path length and lowering the chance of utilizing collisions when appropriate. 
	
	\section{Sampling-based Path Planning Algorithm}\label{sec:Formulation}
	Consider a global (unknown but bounded) map populated with an unknown number of static obstacles. The state $\mathbf{q}=(x,y,\theta,v)$ includes the robot's center of mass position $(x,y)$, as well as the direction $\theta$ and magnitude $v$ of its instantaneous velocity. We assume that 1) the robot is capable of building a local map of the environment from different sensors (such as a LIDAR) as it navigates, 
	2) the obstacles in the environment are closed convex sets, and 
	3) the robot can receive bearing and range information about the goal. 

	 
	

	\subsection{Sampling on Local Configuration Space}
\begin{figure}[!t]
		\vspace{0pt}
		\centering
	\includegraphics[trim={0cm 0.45cm 0.5cm 0.5cm},clip,width=0.80\linewidth]{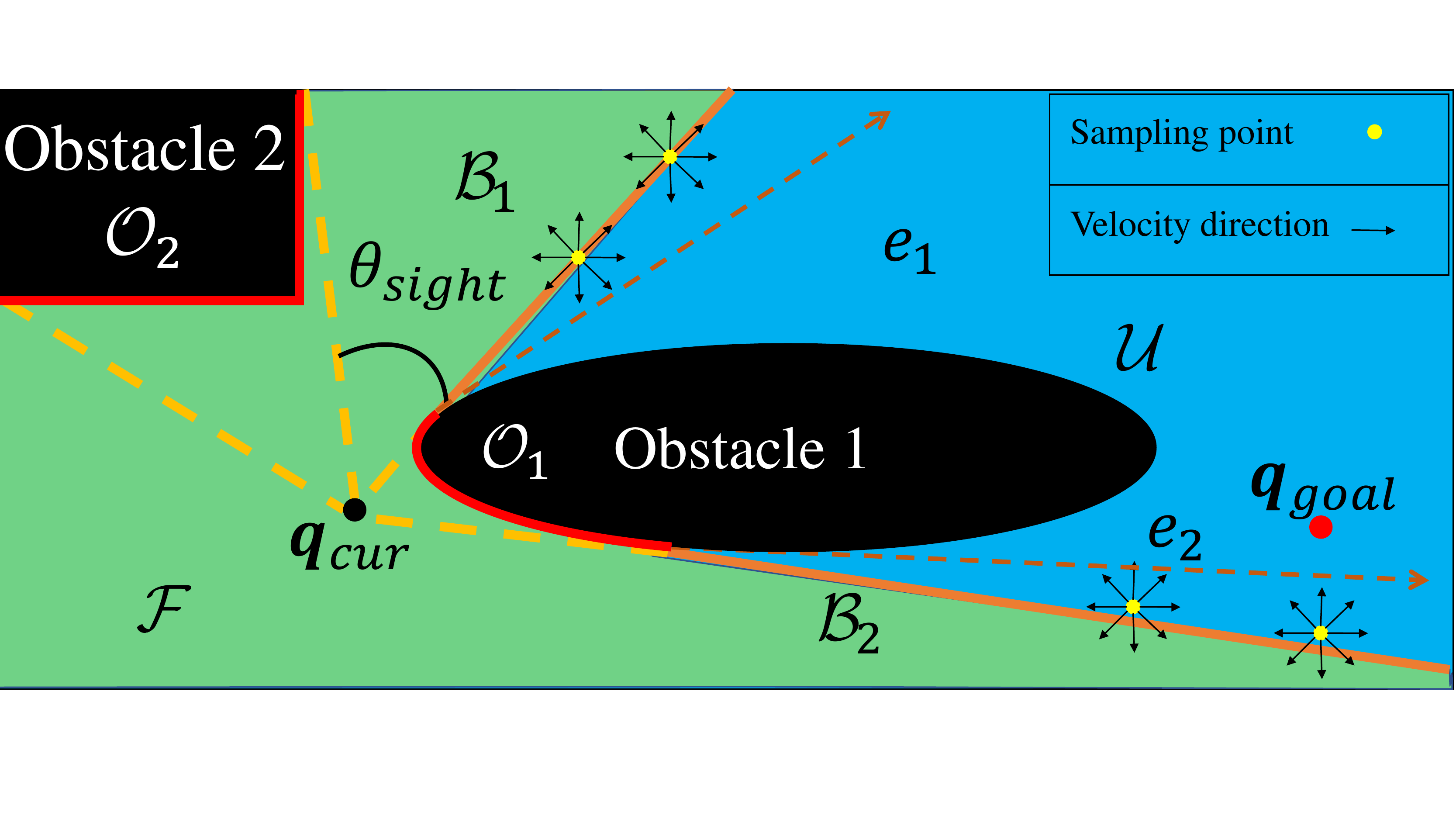}
		\vspace{-12pt}
		\caption{Instance of sampling on a local map with two obstacles.}
		\label{fig:sampling}
		\vspace{-16pt}
	\end{figure}

	Let $\mathcal{M}\subset \mathbb{R}^{2}$ be a sliding local map to 
	represent the environment. The map moves with the robot; its size is a hyper parameter selected by the user. 
	After computing a visibility polygon (e.g., created through ray casting~\cite{bungiu2014efficient}), we extract the local free space $\mathcal{F}$, obstacle-extending frontiers $\mathcal{B}_{i}$, $i=1,\ldots,N$ due to $M$ identified obstacles, and the observable obstacle boundaries $\partial\mathcal{O}_1,\ldots,\partial\mathcal{O}_M$. Then, the local map can be expressed as $\mathcal{M}=\mathcal{F} \cup (\bigcup\limits_{i=1}^{N} \mathcal{B}_{i}) \cup (\bigcup\limits_{i=1}^{M} \partial\mathcal{O}_{i}) \cup \mathcal{U}$, where $\mathcal{U}$ is the unobservable part of the local map (Fig.~\ref{fig:sampling}).\footnote{Note that $U$ contains all partially-observed obstacles with the exception of their observable boundaries.}

	
If the goal is not within $\mathcal{F}$, $\bigcup\limits_{i=1}^{N} \mathcal{B}_{i}$ is sampled to generate a set of candidate states to be explored at future iterations. 
Let $\mathcal{S} = \{s_{\alpha} = (x_{\alpha}, y_{\alpha}): s_{\alpha} \in \bigcup\limits_{i=1}^{N} B_{i}\}$ be the set of positions of all sampled points on the boundary. Positions on each $\mathcal{B}_{i}$ can be sampled according to some user-defined distribution, e.g., uniformly.
Let $\Theta = \{\frac{2\pi n_{l}}{N_{l}}:n_{l}=0,\ldots,N_{l}\}$ be the set of $N_{l} \in \mathbb{N}^{+}$ allowable velocity directions for state expansion. This paper adopts the 8-orientation state expansion, that is $N_{l} = 8$.
	Lastly, let $\mathcal{V} = \{\frac{v_{max}n_{v}}{N_{v}}:n_v=0,\ldots,N_{v}\}$ be the set of $N_{v} \in \mathbb{N}^{+}$ allowable velocity magnitudes for state expansion. 
	We set $N_{v} = \lfloor \frac{length(B_{i})}{\Delta l} \rfloor$, where $\Delta l$ is a hyper parameter that determines sampling on a frontier $\mathcal{B}_{i}$. 
    Then, the sampling configuration space is
	\begin{align*}
	\mathcal{C} = \mathcal{S} \times \Theta \times \mathcal{V} \enspace.
	\end{align*}
	
	The size of $\mathcal{C}$ depends on the size of $\bigcup\limits_{i=1}^{N} \mathcal{B}_{i}$ and how it is sampled, and on the values of $N_{l}$ and $N_{v}$. Sampling the velocity vector space $(\Theta,\mathcal{V})$ presents an interesting trade-off. Larger $N_{l}$ and $N_{v}$ are expected to yield better results overall as they implicitly constrain the position vector space $S$. However, doing so will increase the size of $\mathcal{C}$ and lead to higher computational complexity for planning. Exploring this trade-off is outside the focus of the present paper. 
	While we utilize frontiers, our method is different in that we do not set bounds on velocities based on inevitable collision states but instead evaluate and predict the effect of potential collisions, in unknown environments. Our method can also be tuned to switch between collision exploitation and safety (in the latter case by recovering upper velocity bounds as those used in current collision-avoidance  frontier-based methods). 

	%
The mechanism to switch between collision exploitation and safety (as well as predict the effect of collisions) is based on constructing a special set $\mathcal{E}$.  This set contains the unit vectors tangent to each $\partial\mathcal{O}_j$, $j=1,\ldots,M$ where $\partial\mathcal{O}_j$ and any $B_i$ intersect. For example in Fig.~\ref{fig:sampling}, $\mathcal{E}=\{e_{1}, e_{2} \}$. 
Elements in set $\mathcal{E}$ represent the direction of the \emph{predicted boundaries} of partially-observable obstacles. For convex obstacles, set $\mathcal{E}$ is typically different from the frontier. In the limit that $\mathcal{E}$ and the frontiers coincide, our algorithm will produce a safe collision avoidance behavior as there will be no collisions that can be potentially exploited. (Colliding in that case may be counterproductive.) As we discuss shortly, predicted boundaries are critical in order to be able to estimate possible future collisions, and serve as an integral part in defining this work's risk (in terms of obstacle collision) and collision reward cost functions.

	\subsection{Formulating the Planner Optimization Function}
	
	The planner seeks to select a sequence of intermediate states $\mathbf{q}_{s} \in \mathcal{C}$ that take the robot from a current state $\mathbf{q}_{cur}$ (in the local map) to the goal state $\mathbf{q}_{goal}$. 
	%
	Each intermediate state $\mathbf{q}_{s} \in \mathcal{C}$ is calculated 
	by solving the (local) unconstrained optimization problem 
	%
    \begin{equation}\label{eq:optimization}
	\mathbf{q}_{s} = \mathop{\arg\min}_{\mathbf{q}_{s} \in \mathcal{C}} \left[ w_{p}J_{pos} + w_{r}J_{risk} + w_{v}J_{vel} \right] \enspace.
	\end{equation}
	%
	The cost function in~\eqref{eq:optimization} is a weighted sum of individual costs $\left\{J_{pos},J_{risk},J_{vel}\right\}$, each representing distinct objectives:
	\begin{itemize}
    \item Minimizing $J_{pos}$ finds the shortest path between $\mathbf{q}_{cur}$ and $\mathbf{q}_{goal}$. (See Section~\ref{sss:Jpos} and Algorithm~\ref{pseudo J_pos}.)
    \item Minimizing $J_{risk}$ chooses the least risky path in terms of avoiding collisions. (See Section~\ref{sss:Jrisk} and Algorithm~\ref{pseudo J_risk}.)
    \item Minimizing $J_{vel}$ picks an as high as possible number of beneficial collisions while maintaining as high speed as possible; \emph{this is where the novelty of our cost function arises from}. Instead of designing a stop maneuver to ensure the robot always stays in free known space, this cost will generate a potential to make the robot go and exploit the unknown area, colliding with obstacles and using collisions as a means to steer toward the goal. (See Section~\ref{sss:Jvel}.) 
    \end{itemize}
	%
	%
	%

Values for weights $\left\{w_{p}, w_{r}, w_{v}\right\}$ are hyper parameters selected by the user, and allow the algorithm to be tuned for balancing safety, risk, and exploration. 
    

\subsubsection{Generating $J_{pos}$}\label{sss:Jpos}
To determine an intermediate state $\mathbf{q}_{s}$, we begin by computing $J_{pos}$ via Algorithm~\ref{pseudo J_pos}. In the inputs to Algorithm~\ref{pseudo J_pos}, $\mathbf{p}_{cur}$ and $\mathbf{p}_{goal}$ are the current and goal positions, $\mathbf{p}_{s} \in \mathcal{S}$ is the position of the candidate point in the sampling configuration space, 
$\mathbf{p}_{pre}$ is the previous position, and $\mathbf{p}_{ini}$ is the initial starting point. 
	%
	%
A penalty factor $f_{a} \in [1,\infty]$ is applied to force the robot not to vary its orientation beyond a prescribed threshold $\theta_{thres}$. 
	
We first approximate the length of the path from the current position $\mathbf{p}_{cur}$ to the goal position $\mathbf{p}_{goal}$ which consists of the actual cost $PathLength(\mathbf{p}_{ini}, \mathbf{p}_{pre}, \mathbf{p}_{cur}, \mathbf{p}_{s})$ and the heuristic cost $PathLength(\mathbf{p}_{s}, \mathbf{p}_{goal})$. We then assign the penalty factor $f_{a}$ to regulate direction of motion, and normalize the output to yield $J_{pos} = normalize(P_{pose})$.


	\begin{algorithm}[h!]
		\caption{Generate $P_{pos}$}
		\label{pseudo J_pos}
		\LinesNumbered
		\SetKwInOut{Input}{input}
		\SetKwInOut{Output}{output}
		\Input{$\mathbf{p}_{cur}$, $\mathbf{p}_{goal}$, $\mathbf{p}_{s}$, $\mathbf{p}_{pre}$, $\mathbf{p}_{ini}$, $\theta_{thres}$, $f_{a}$}
		\Output{$P_{pos}$}
		    $PL_{actual} \leftarrow PathLength(\mathbf{p}_{ini}, \mathbf{p}_{pre}, \mathbf{p}_{cur}, \mathbf{p}_{s})$\\
		    $PL_{heuristic} \leftarrow PathLength(\mathbf{p}_{s}, \mathbf{p}_{goal})$\\
			$D_{pos} \leftarrow PL_{actual} + PL_{heuristic}$\\ 
			\eIf {$\angle(\mathbf{p}_{pre}, \mathbf{p}_{current}, \mathbf{p}_{s}) \geq \theta_{thres}$}
			{$P_{pos} \leftarrow f_{a} \times D_{pos}$ \\
    			}{$P_{pos} \leftarrow D_{pos}$}
		\Return $P_{pos}$
	\end{algorithm}
	
	

	\subsubsection{Generating $J_{risk}$ for a point}\label{sss:Jrisk}
	We define the risk as the probability of colliding with obstacles. To generate $J_{risk}$, which is an explicit evaluation of probability of collision, we follow Algorithm~\ref{pseudo J_risk}. 
	In the inputs to Algorithm~\ref{pseudo J_risk}, $\mathbf{v}_{s}$ is the velocity of the candidate point, $\mathcal{E}$ is the set of unit vectors $e_{i}$ of predicted obstacle boundaries, $\mathbf{p}_{p} = (x_{p}, y_{p})$ is the potential collision point based on $\mathcal{E}$, and $\delta v$ is a positive infinitesimal value used to evaluate $J_{risk}$ based on the distance between the robot and the obstacles when the robot is stationary. 
	
	\begin{algorithm}[h!]
		\caption{Generate $J_{risk}$}
		\label{pseudo J_risk}
		\LinesNumbered
		\SetKwInOut{Input}{input}
		\SetKwInOut{Output}{output}
		\Input{$\mathbf{p}_{s}$, $\mathbf{v}_{s}$, $\mathbf{p}_{p}$, $\mathcal{E}$, $\delta v$}
		\Output{$J_{risk}$}
		\eIf{$\lVert \mathbf{v}_{s}\rvert > 0$}{
			$\mathbf{s}_{p}\leftarrow \mathbf{p}_{p} - \mathbf{p}_{s}$\\
			$d_{p} \leftarrow \lVert \mathbf{s}_{p} \rVert$\\
			$v_{c} \leftarrow \lVert Proj(\mathbf{v}_{s}, \mathbf{s}_{p})\rVert$
		}{$v_{c}\leftarrow \delta v$\\
			$d_{p} \leftarrow \mathop{\arg\min}_{\mathbf{e}_{i}\in \mathcal{E}}{distance(\mathbf{p}_{s}, \mathbf{e}_{i})}$}
		$t_{c} \leftarrow \frac{d_{p}}{v_{c}}$\\
		$J_{risk,h} \leftarrow normalize(\frac{1}{t_{c}})$\\
		$J_{risk, g} \leftarrow  -f_{\theta}\theta_{sight} $ \\
		$J_{risk} \leftarrow J_{risk, g} + J_{risk,h}$\\
		\Return $J_{risk}$
	\end{algorithm}

	We determine the heuristic part of $J_{risk}$ by calculating the collision time $t_{c}$ \cite{song2018t} based on the current state and the predicted boundary of the obstacle in $\mathcal{U}$ if the candidate velocity $\lVert\mathbf{v}_{s} \rVert> 0$. Otherwise, we  generate $J_{risk}$ given $\delta v$ and ${d}_{p}$ where $d_{p}$ is the minimal distance between the robot and the obstacles in the environment. 
	We use $J_{risk, g} = -f_{\theta}\theta_{sight}$ (line 11), where $\theta_{sight}$ is the angle of the narrow region as illustrated in Fig.~\ref{fig:sampling}. Weighting $\theta_{sight}$ with a constant penalizing factor $f_{\theta} \in [0,1]$ helps prevent the robot from going into narrow regions in $\mathcal{F}$. 

\subsubsection{Generating $J_{vel}$}\label{sss:Jvel}
	%
This cost forces the robot to exploit the unknown area, colliding with obstacles and using collisions as a means to steer toward the goal. It is the distinctive element with respect to other related algorithms (e.g.,~\cite{tordesillas2019real,lopez2017aggressive, oleynikova2018safe}) in the sense that it relaxes the condition of calculating and enforcing inevitable collision states (ICS). 
In \cite{StagerISER18}, a reflection model is generated to fit the behavior of a holonomic robot after colliding with the environment. Works~\cite{StagerISER18} and~\cite{mulgaonkar2017robust} show that bouncing motion primitives can be leveraged in fast planning algorithms to generate minimum-time trajectories. We simplify the complex problem of vehicle-surface interactions using Garwins model~\cite{garwin1969kinematics}. Then we can generate an explicit reward function to evaluate whether the robot can benefit from the reflection-like bouncing behavior.
	%
	%
    \begin{equation}\label{eq:ref}
    \mathbf{R}_{ref}(t)=\left\{ \begin{array}{rcl}
    \langle \mathbf{v}_{ref},\mathbf{unit}_{goal}\rangle & & {\mathbf{p}_{ref} \notin \mathcal{F}}\\
    penalty & & {otherwise} \enspace.
    \end{array} \right. 
    \end{equation}

    	\begin{figure*}[!t]
	\vspace{3pt}
	\centering
	\begin{subfigure}[b]{0.32\textwidth}
		\includegraphics[trim={5cm 0.5cm 5cm 0.5cm},clip,width=0.98\linewidth]{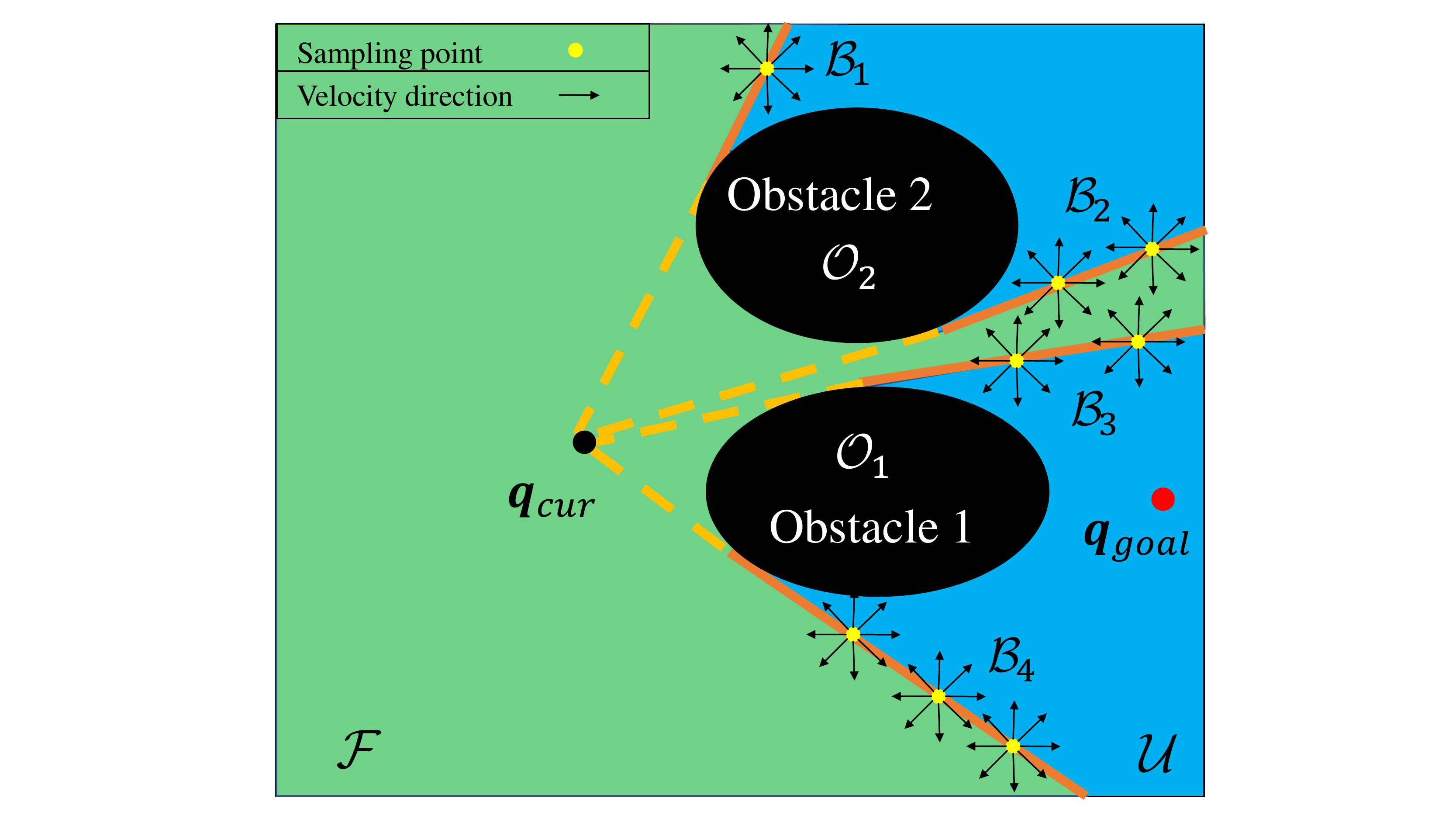}
		\vspace{-15pt}
		\caption{Before pruning}
		\label{fig:before pruning}
	\end{subfigure}
	\begin{subfigure}[b]{0.32\textwidth}
		\includegraphics[trim={5cm 0.5cm 5cm 0.5cm},clip,width=0.98\linewidth]{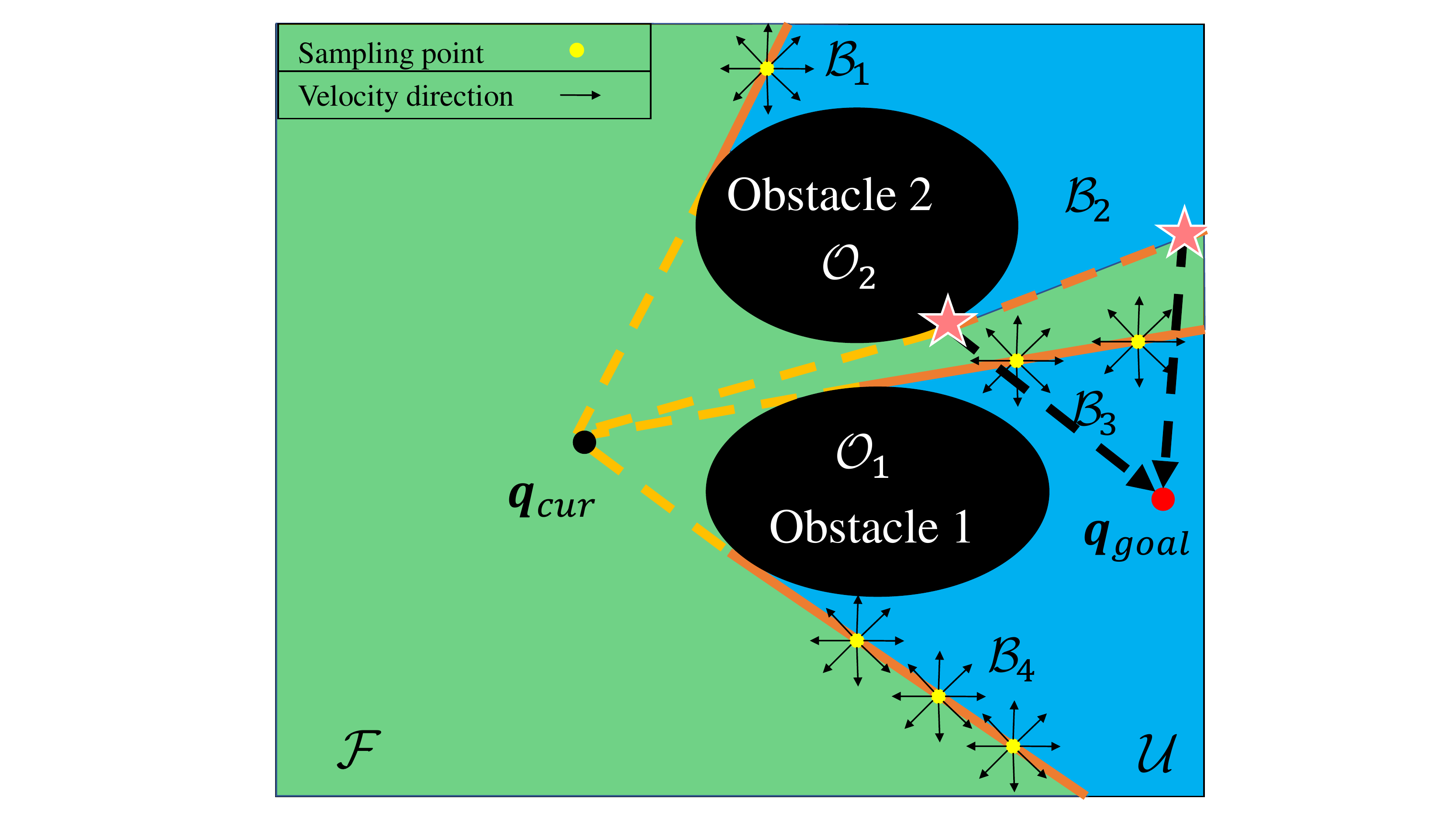}
		\vspace{-15pt}
		\caption{Position pruning}
		\label{fig:pos pruning}
	\end{subfigure}
	\begin{subfigure}[b]{0.32\textwidth}
		\includegraphics[trim={5cm 0.5cm 5cm 0.5cm},clip,width=0.98\linewidth]{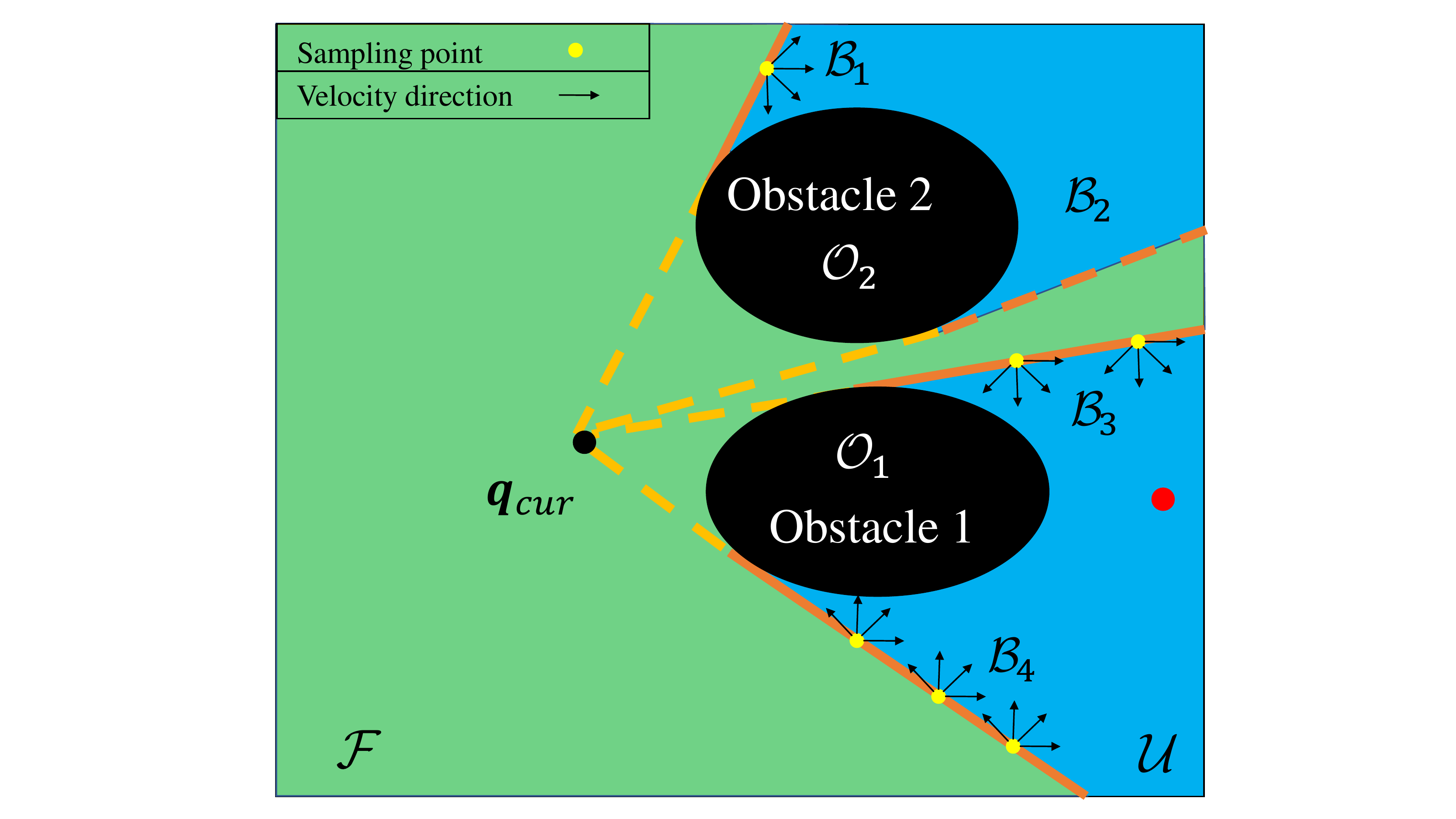}
		\vspace{-15pt}
		\caption{Velocity pruning}
		\label{fig:vel pruning}
	\end{subfigure}
	\vspace{-3pt}
	\caption{Example of pruning on the sampling configuration space $\mathcal{C}$.}
	\vspace{-9pt}
	\label{fig:pruning}
\end{figure*}

    	The reflection velocity $\mathbf{v}_{ref}$ in~\eqref{eq:ref} is the instantaneous velocity of the robot after impact. It is calculated by taking the norm of $\mathbf{e}_{p}$ which lies along the predicted boundary and is directed toward $\mathcal{U}$. We apply a rotation to $\mathbf{e}_{p}$ and denote the normal vector as $\mathbf{n}_{p} = Rot(\mathbf{e}_{p}, \frac{\pi}{2})$. $\mathbf{v}_{ref}$ can then be derived from $\mathbf{v}_{s}$ and $\mathbf{n}_{p}$ based on the reflection model. $\mathbf{p}_{ref}$ is an estimation of the robot's position after a user-defined time interval. $penalty$ is a penalty value in $[-\infty, 0]$ to prevent the robot bouncing back to the currently known space. %

	
	After we compute the explicit heuristic function to evaluate risk and reflection in unknown space, we then generate $J_{vel}$ which represents how the impact velocity on the boundary will affect future behavior of the robot. The effect of velocity in $\mathcal{C}$ can then be approximated by  
	%
	\begin{equation}\label{eq:vel}
	J_{vel} = -\langle\mathbf{v}_{s}, \mathbf{p}_{goal} - \mathbf{p}_{cur}\rangle (1-J_{risk, h}) - R_{ref}J_{risk, h}\enspace.
	\end{equation}

\subsection{Candidate State Pruning for Online Implementation}\label{ss:pruning}
To make the algorithm online, we develop a two-step pruning technique for reducing the number of candidate states $\mathbf{q}_s \in \mathcal{C}$. States are removed based on either their position and/or their velocity vector (see Fig.~\ref{fig:pruning}).

	
\vspace{2pt}
\subsubsection*{\bf Position-Based Pruning} In position-based pruning, candidate states with positions 
that are less likely to be selected are dynamically identified and removed from $\mathcal{C}$. 
The chance for a candidate state to be selected based on its position alone depends on the number of neighboring candidate states and their relative distance to the goal, and whether there is separation by free or unobservable space. For example, in the illustrations shown in Fig.~\ref{fig:before pruning} and Fig.~\ref{fig:pos pruning}, candidate states lying on $\mathcal{B}_2$ are pruned because they are close to candidate states lying on $\mathcal{B}_3$ which in turn are closer to the goal, and they are separated from them by free space $\mathcal{F}$.

Position-based pruning does not remove candidate states that may be far from the goal (such as those lying on $\mathcal{B}_1$ and $\mathcal{B}_4$) to promote exploration and to handle potential uncertainties (e.g., robot drift or under-estimation of collision impact). Instead, it removes states that are expected to have a similar effect, if selected, toward exploration. 
The completeness can be guaranteed assuming that the obstacle is convex, since the algorithm always enforces exploration of the unknown space $\mathcal{U}$. However, the pruning technique cannot guarantee completeness in maze-like environments.


\vspace{2pt}
\subsubsection*{\bf Velocity-Direction-Based Pruning} 
In velocity-based pruning, candidate states with velocities directed toward the free space $\mathcal{F}$ are pruned. As in position-based pruning, the objective here is to retain the candidate states with velocities that encourage exploration. 
For example, in the illustrations shown in Fig.~\ref{fig:pos pruning} and Fig.~\ref{fig:vel pruning}, all candidate states remaining after position-based pruning are evaluated once more, and only those that lead toward unobservable space $\mathcal{U}$ are kept.


	
    \section{Trajectory Generation}\label{sec:Algorithm}
	
	We are now ready to introduce our trajectory generation algorithm for collision-resilient robots navigating in unknown obstacle-cluttered environments. Recall that no prior knowledge of the map is required. 
	The algorithm consists of two components: 1) generating appropriate waypoints via B-spline interpolation, and 2) selecting appropriate low level collision-free or collision-harnessing maneuvers. 


\begin{figure*}[!t]
	\vspace{3pt}
	\centering
	\begin{subfigure}[b]{0.32\textwidth}
		\includegraphics[trim={5cm 0.4cm 5cm 0.6cm},clip, width=0.98\linewidth]{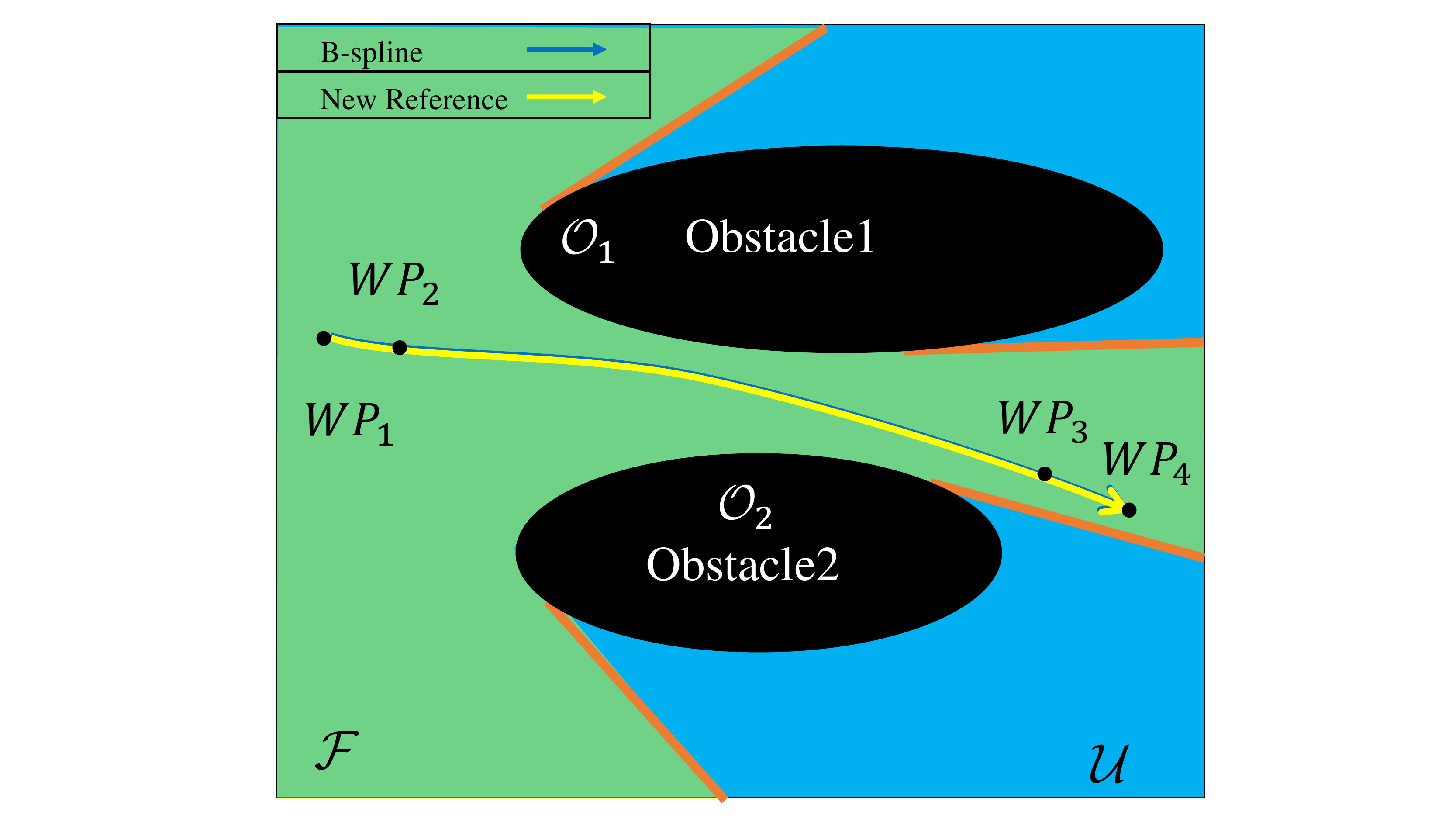}
		\vspace{-3pt}
		\caption{Free space}
		\label{fig:free space}
	\end{subfigure}
	\begin{subfigure}[b]{0.32\textwidth}
		\includegraphics[trim={5cm 0.5cm 5cm 0.5cm},clip, width=0.98\linewidth]{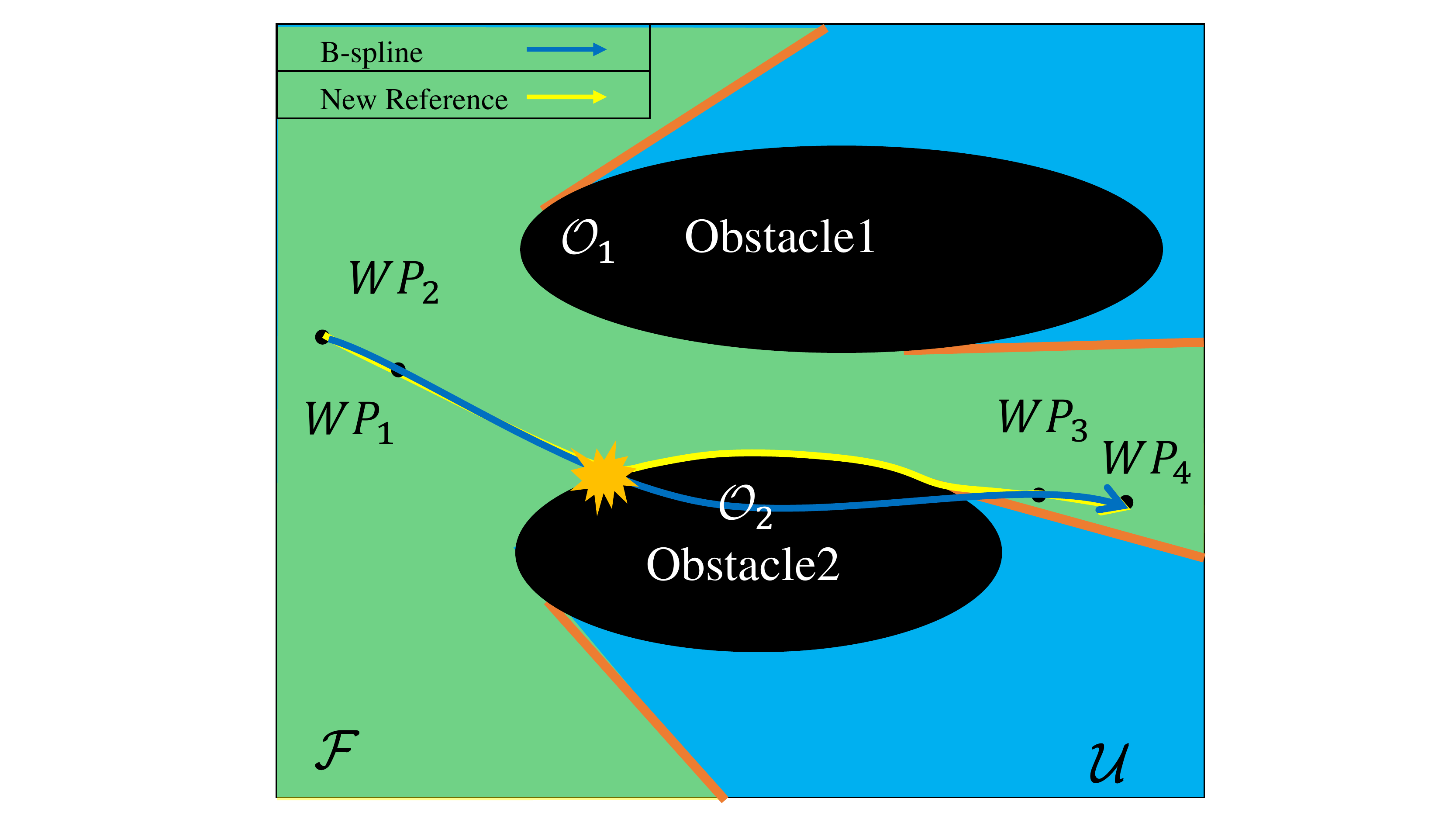}
		\vspace{-3pt}
		\caption{Boundary following}
		\label{fig:boundary following}
	\end{subfigure}
	\begin{subfigure}[b]{0.32\textwidth}
		\includegraphics[trim={5cm 0.5cm 5cm 0.5cm},clip,width=0.98\linewidth]{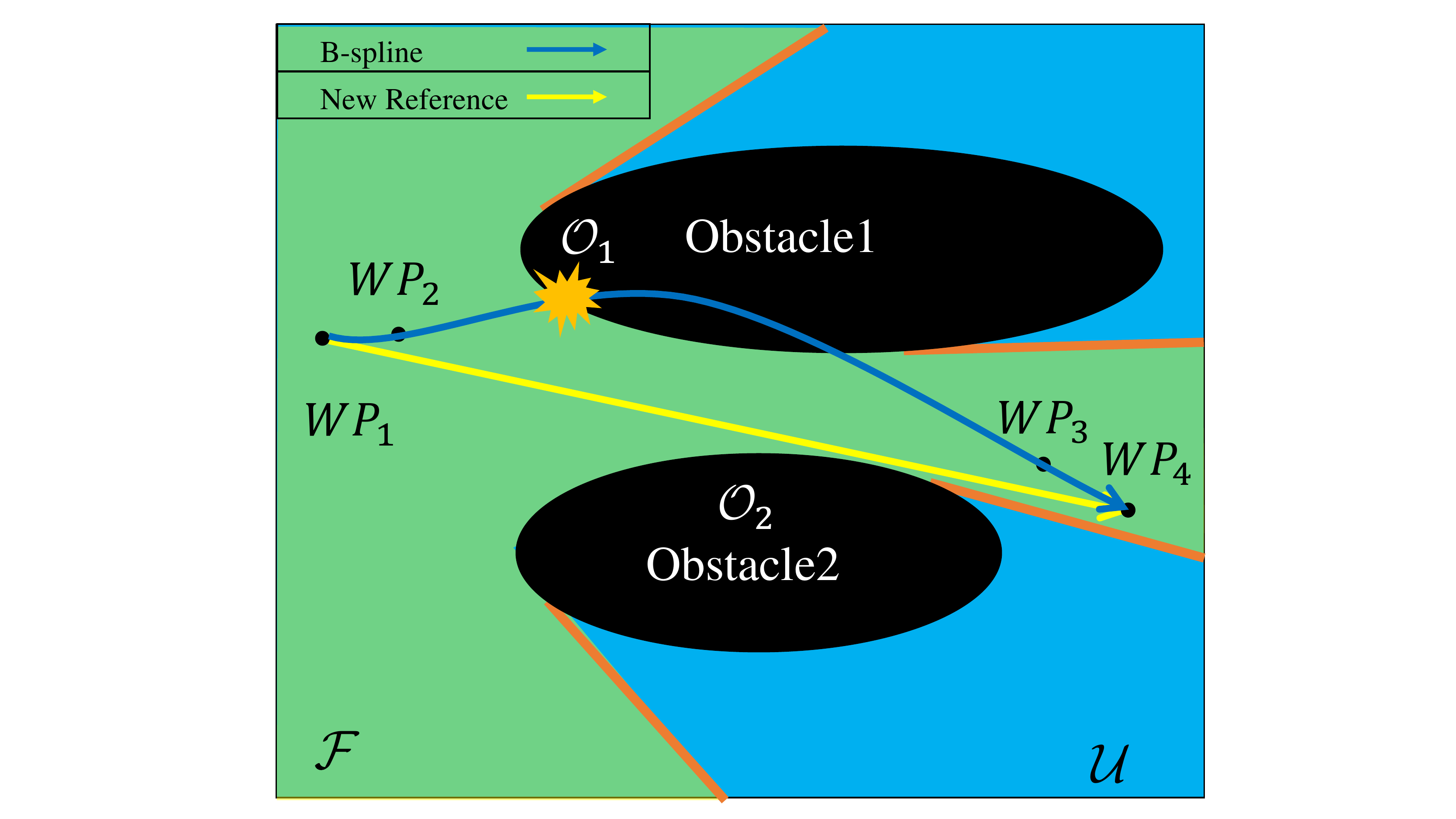}
		\vspace{-3pt}
		\caption{Flow-through}
		\label{fig:flow-through}
	\end{subfigure}
	\vspace{-3pt}
	\caption{The switching controller deals with potential collisions that may occur in future time.}
	\vspace{-12pt}
	\label{fig:controller}
\end{figure*}

	
	\subsection{B-spline Generation}

	Once the intermediate state $\mathbf{q}_{s}$ is found, we use the cubic B-spline~\cite{elbanhawi2015continuous} to interpolate a spline with control points as $WP$. The Python B-spline interpolation function in the scipy package is used to generate a cubic B-spline $Spl$ is the set of points in the B-spline. A trajectory is then generated from $Spl$ using a time interval $T$ defined as
\begin{equation}
T = \max\{T_{v}, T_{map} \}\enspace, 
\end{equation} 
where $T_{map}$ is the time needed to update the sliding map and $T_{v}$ is the maximum velocity of the robot. We can approximate $T_{v}$ by
\begin{equation}
 T_{v} = P_{safe} \times \frac{\sum\limits_{i=2}^{N} \lVert WP[i] - WP[i-1]\rVert}{v_{max}}
\end{equation}
where $P_{safe}$ is the parameter which restricts the reference velocity to stay within the range $[0, v_{max}]$. In our case, we choose $N =4$ to keep the minimal number of control points for cubic spline.


\subsection{Maneuver Selection}
We utilize a low-level switching controller to ensure that the robot follows the trajectory and reaches the desired waypoint. At each instance when the sliding local map is updated, the controller switches between the following strategies: free-space, boundary following~\cite{stager2016stochastic,lu2019optimal}, and flow-through~\cite{StagerICRA19} as described in Algorithm~\ref{pseudo control_switch}. From the switching strategy, the robot will determine whether the B-spline trajectory is suitable or it needs to generate a new reference trajectory as illustrated in Fig.~\ref{fig:controller}. In all strategies, the robot follows reference trajectories using a PID controller.

\begin{algorithm}[h!]
		\caption{Maneuver Selection}
		\label{pseudo control_switch}
		\LinesNumbered
		\SetKwInOut{Input}{input}
		\SetKwInOut{Output}{output}
		\Input{$\mathbf{WP}$, $\mathbf{p}_{c}$ }
		\Output{$controller\_class$}
		\eIf{$\mathbf{p}_{c} = \emptyset$}{$controller\ class = free\_space$}{$\mathbf{v}\leftarrow \frac{\mathbf{WP}_{4} - \mathbf{WP}_{1}}{\lVert\mathbf{WP}_{4} - \mathbf{WP}_{1}\rVert}$\\
		$\mathbf{v}\leftarrow Rot(\mathbf{v},\frac{\pi}{2})$\\
		$\mathbf{v}_{c}\leftarrow \mathbf{p}_{c} - \mathbf{WP}_{1}$\\
		\eIf{$\langle \mathbf{v},\mathbf{v}_{c}\rangle \geq 0$}{$controller\_class = flow\_through$}{$controller\_class = boundary\_following$}}
		\Return $controller\_class$
\end{algorithm}

When the robot does not face a collision, it utilizes the free-space strategy and follows the trajectory generated from the B-spline $\mathbf{q}_{spl}$ as depicted in Fig.~\ref{fig:controller}a. 
If colliding at position $\mathbf{p}_{c}$, the robot will utilize either the flow-through or the boundary following strategy. 
Let $\mathbf{q}_{bound}$ be the trajectory projection on the boundary. 
If $\langle \mathbf{v},\mathbf{v}_{c}\rangle < 0$, then the robot will use the boundary following strategy. The collision point $\mathbf{p}_{c}$, where the B-spline $\mathbf{q}_{spl}$ first intersects the obstacle boundary $\mathbf{b}_{\mathcal{O}}$, is referred to as the engaging point $\mathbf{p}_{d}$. At this point, the robot will follow a newly generated reference trajectory $\mathbf{q}_{bound}$, generated by projecting $\mathbf{q}_{spl}$ to the boundary $\mathbf{b}_{\mathcal{O}}$ up until the disengaging point. The boundary following maneuver is depicted in Fig.~\ref{fig:controller}b.
\begin{equation*}
\mathbf{q}_{ref}(t)=\left\{ \begin{array}{rcl}
\mathbf{q}_{spl}(t) & & {s = off}\\
\mathbf{q}_{bound}(t) & & {s = on} \enspace.
\end{array} \right. 
\end{equation*} 
\begin{equation*}
\phi(s, x) =\left\{ \begin{array}{rcl}
off && {s = off,\ \lVert \mathbf{p}(t) - \mathbf{p}_{c}\rVert \geq \delta}\\
on && {s = off,\ \lVert \mathbf{p}(t) - \mathbf{p}_{c}\rVert \leq \delta}\\
off && {s = on,\ \lVert \mathbf{p}(t) - \mathbf{p}_{d}\rVert \leq \delta}
\enspace.
\end{array} \right. 
\end{equation*}

If $\langle \mathbf{v},\mathbf{v}_{c}\rangle \geq 0$ and ${q}_{spl} \cap \mathcal{O}\neq \emptyset$, then the robot will use the flow-through maneuver as depicted in Fig.~\ref{fig:controller}c. $\mathbf{q}_{flow}$ is generated by projecting $\mathbf{q}_{spl}$ to vector $WP[4] - WP[1]$. 
\begin{equation*}
\mathbf{q}_{ref}(t)=\left\{ \begin{array}{rcl}
\mathbf{q}_{spl}(t) & & {s = off}\\
\mathbf{q}_{flow}(t) & & {s = on} \enspace.
\end{array} \right. 
\end{equation*} 
\begin{equation*}
\phi(s, x) =\left\{ \begin{array}{rcl}
off && {s = off,\ \lVert \mathbf{p}(t) - \mathbf{p}_{c}\rVert \geq \delta}\\
on && {s = off,\ \lVert \mathbf{p}(t) - \mathbf{p}_{c}\rVert \leq \delta}
\enspace.
\end{array} \right. 
\end{equation*}


%


\begin{figure*}[!t]
	\vspace{0pt}
	\centering
	\begin{subfigure}[b]{0.325\textwidth}
		\includegraphics[width=1.0\linewidth]{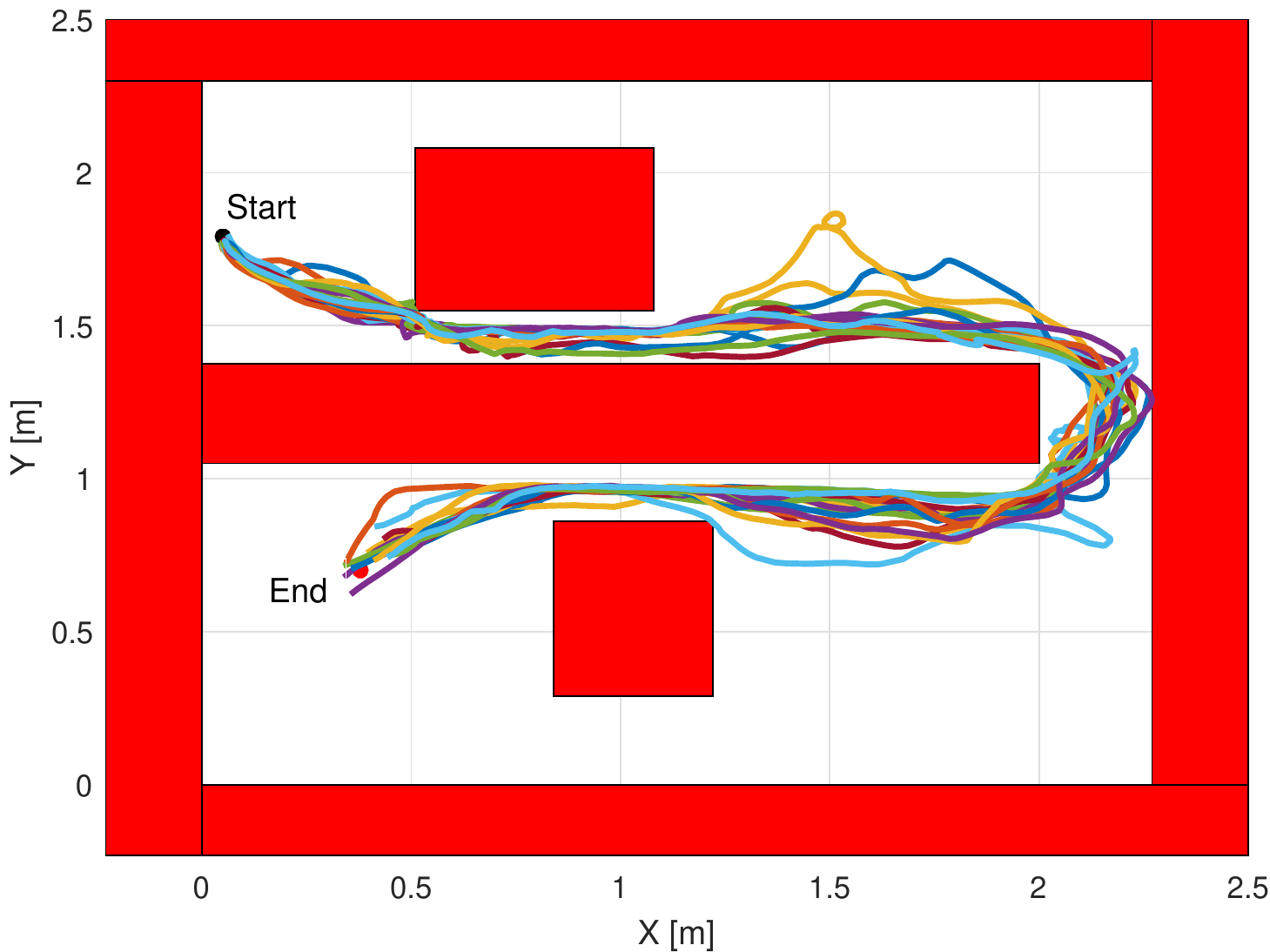}
		\vspace{-15pt}
		\caption{$w_{r}=0.1,\ w_{v}=4.0$}
		\label{fig:refleciton}
	\end{subfigure}
	\begin{subfigure}[b]{0.325\textwidth}
		\includegraphics[width=1.0\linewidth]{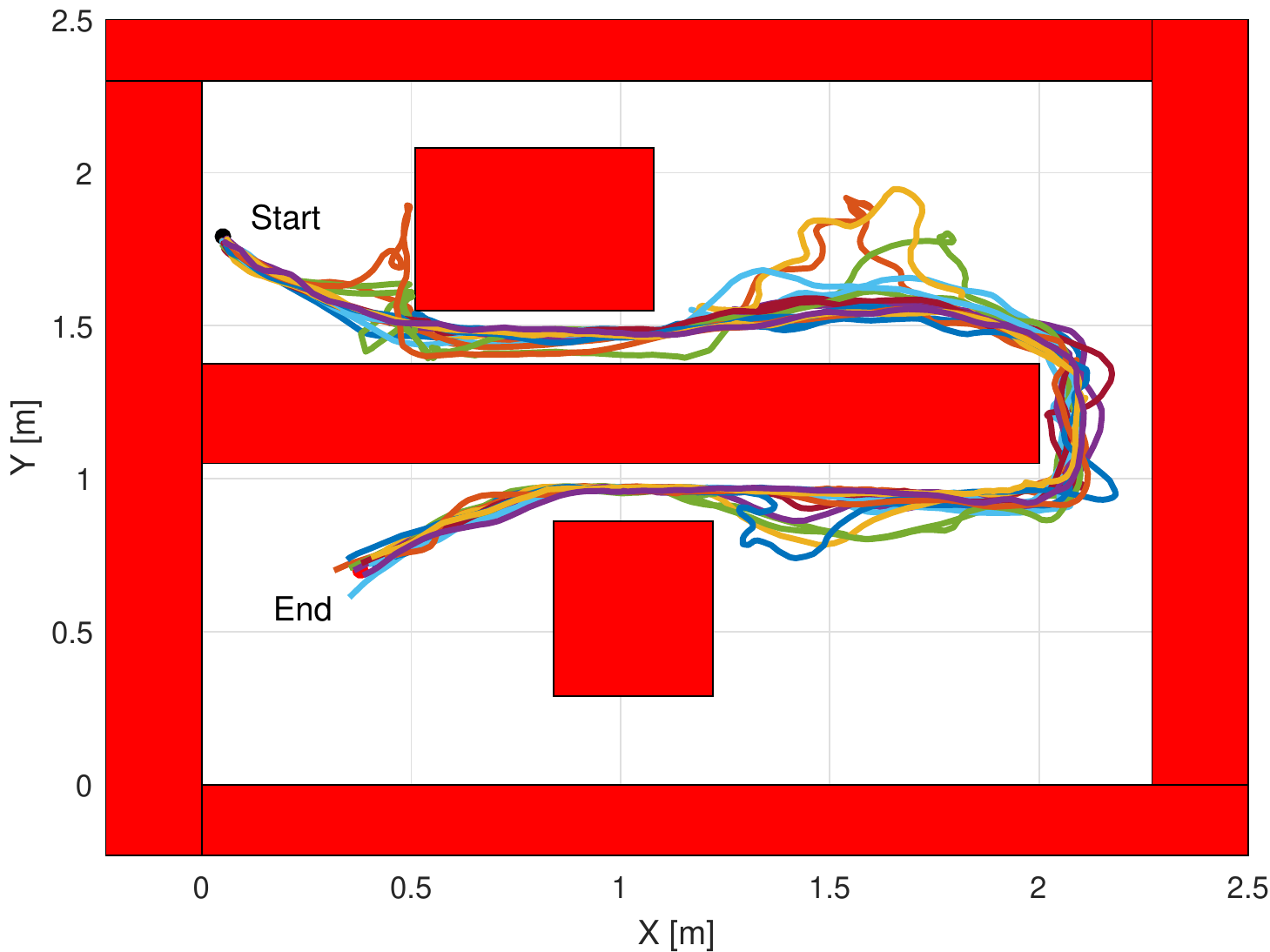}
		\vspace{-15pt}
		\caption{$w_{r}=0.1,\ w_{v}=0.0$}
		\label{fig:high risk}
	\end{subfigure}
	\begin{subfigure}[b]{0.325\textwidth}
		\includegraphics[width=1.0\linewidth]{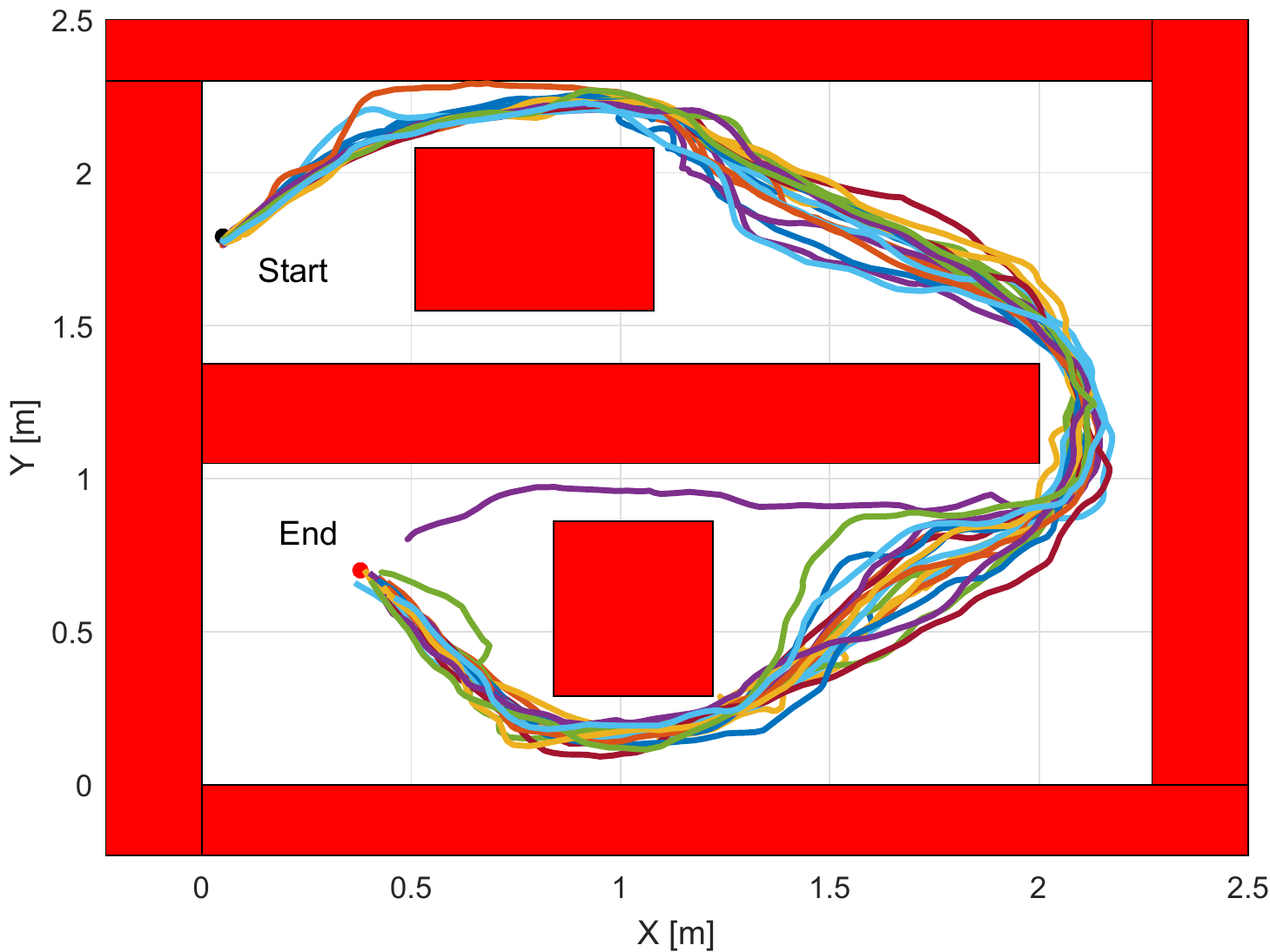}
		\vspace{-15pt}
		\caption{$w_{r}=100.0,\ w_{v}=0.0$}
		\label{fig:low risk}
	\end{subfigure}
	\vspace{-1pt}
	\caption{Experimental trajectories with $T_{map} = 0.2s$. (a) High risk path with intentional collision reflection. (b) High risk path without intentional collision reflection. (c) Low risk path without intentional collision reflection.}
	\vspace{-12pt}
	\label{fig:example three cases}
\end{figure*}

\section{Experimental Results}\label{sec:Experiment}

\vspace{0pt}
\subsection{Robot and Environment Setup}
We implement our proposed algorithm on the collision-resilient Omnipuck platform~\cite{StagerISER18} which we built in-house (Fig.~\ref{fig:introFIG}). The body of the Omnipuck robot is surrounded by a reflection ring that enables it to collide safely with the environment and rebounce from it. The robot operates in an $2.4m \times 2.4m$ confined corridor environment with rectangular pillars serving as static convex obstacles as shown in Fig.~\ref{fig:introFIG}. 
The position of Omnipuck is captured using a 12-camera VICON motion capture system. A laptop with a $2.3$\;GHz i7 CPU and $12$\;GB RAM processes position data and sends controller commands to the robot at a frequency of $20$\;Hz. We utilize the ray casting algorithm~\cite{bungiu2014efficient} to simulate LIDAR behavior and generate a current sliding map. See \protect\url{https://www.youtube.com/watch?v=S3oYebJRfA0} for instances of our experiments. 

\vspace{-3pt}
\subsection{Effect of Harnessing Collisions}\label{subsect:Harnessing}
To examine the effect of different weight factors $\mathbf{w}=[w_{p},w_{r},w_{v}]$ in~\eqref{eq:optimization} on planning the motion of a collision-resilient robot in unknown environments, we select weight factors that emphasize components of the objective function ~\eqref{eq:optimization} to generate either a low/high risk path or a path that harnesses collisions. 
We consider three case studies as follows. 
\begin{itemize}
    \item A high risk trajectory that intentionally collides with the environment to harness collisions is generated when~\eqref{eq:optimization} is optimized with $w_{v}$ set to a large value.
    \item A high risk trajectory that may unintentionally collide with the environment without harnessing collisions is generated when~\eqref{eq:optimization} is optimized with $w_{r}$ set to a small value and $w_{v}$ set to zero.
    \item  A low risk (`safe') trajectory that seeks to avoid collisions altogether is generated when~\eqref{eq:optimization} is optimized with $w_{r}$ set to a large value and $w_{v}$ set to zero.
\end{itemize}

A higher $w_{v}$ is expected to produce a risky trajectory which can utilize the reflection after colliding to continue making progress toward the goal. We anticipate that in this way the robot can reach the goal with a shorter arrival time and path length. To test this hypothesis, we run $20$ closed-loop experimental trials with map update time $T_{map} = 0.2$\;sec with each of the three strategies listed above. The top speed in all cases is ${v}_{max}=1.2\ m/s$.

Experimental results are shown in Fig.~\ref{fig:example three cases}. Individual trials for each case are overlaid on an augmented-obstacle representation of the environment map. We notice that when the planner can tolerate more risk (i.e. lower $w_r$), then the robot will try go through the two narrow corridors to decrease arrival time and total path length.  However, if no mechanism to harness collisions is in place, and unintentional collisions do happen, some of them may have a negative effect on the robot trajectory. For example, in Fig.~\ref{fig:example three cases}(b) some trials collide with the left side of the top obstacle, effectively increasing both the arrival time and the total path length. On the contrary, if the planner deliberately seeks to collide when needed (Fig.~\ref{fig:example three cases}(a)) we observe that no such negative collisions occur. As expected, a risk-averse (i.e. higher $w_r$) planner will seek for longer trajectories through the wider parts of the environment (Fig.~\ref{fig:example three cases}(c)).


The improvement ($\%$) when utilizing collisions in terms of arrival times and total path length can be clearly seen in Table~\ref{table:improvement}. Results reveal that our proposed planning algorithm to harness collisions has a major impact in terms of mean arrival time. Mean total path length is similar to the case of high-risk planning without explicitly harnessing collisions, but significantly better that in the case of low-risk planning. 

\begin{table}[!h]
\vspace{0pt}
    \caption{Improvements of collision-harnessing planning in terms of arrival time and total path length when compared to high-risk and low-risk planning without intentional collisions.}
    \vspace{-6pt}
    \label{table:improvement}
    \begin{center}
    \begin{tabular}{|c|c|c|}
    \hline
    \textbf{Improvement} ($\%$) & \textbf{High risk path} & \textbf{Low risk path}\\
    \hline
    Arrival time mean 
    & $7.60$
    & $8.53$\\
    \hline
    Arrival time STD 
    & $3.48$
    & $2.39$\\
    \hline
    Path length mean
    & $2.32$
    & $16.01$\\
    \hline
    Path length STD 
    & $0.99$
    & $0.47$\\
    \hline
    \end{tabular}
    \end{center}
    \vspace{-8pt}
\end{table}


%

\vspace{-6pt}
\subsection{Parametric Study on Different Mapping Times}

\begin{figure}[h!]
\vspace{3pt}
 \begin{subfigure}{.235\textwidth}
  \centering
  \includegraphics[width=0.95\linewidth]{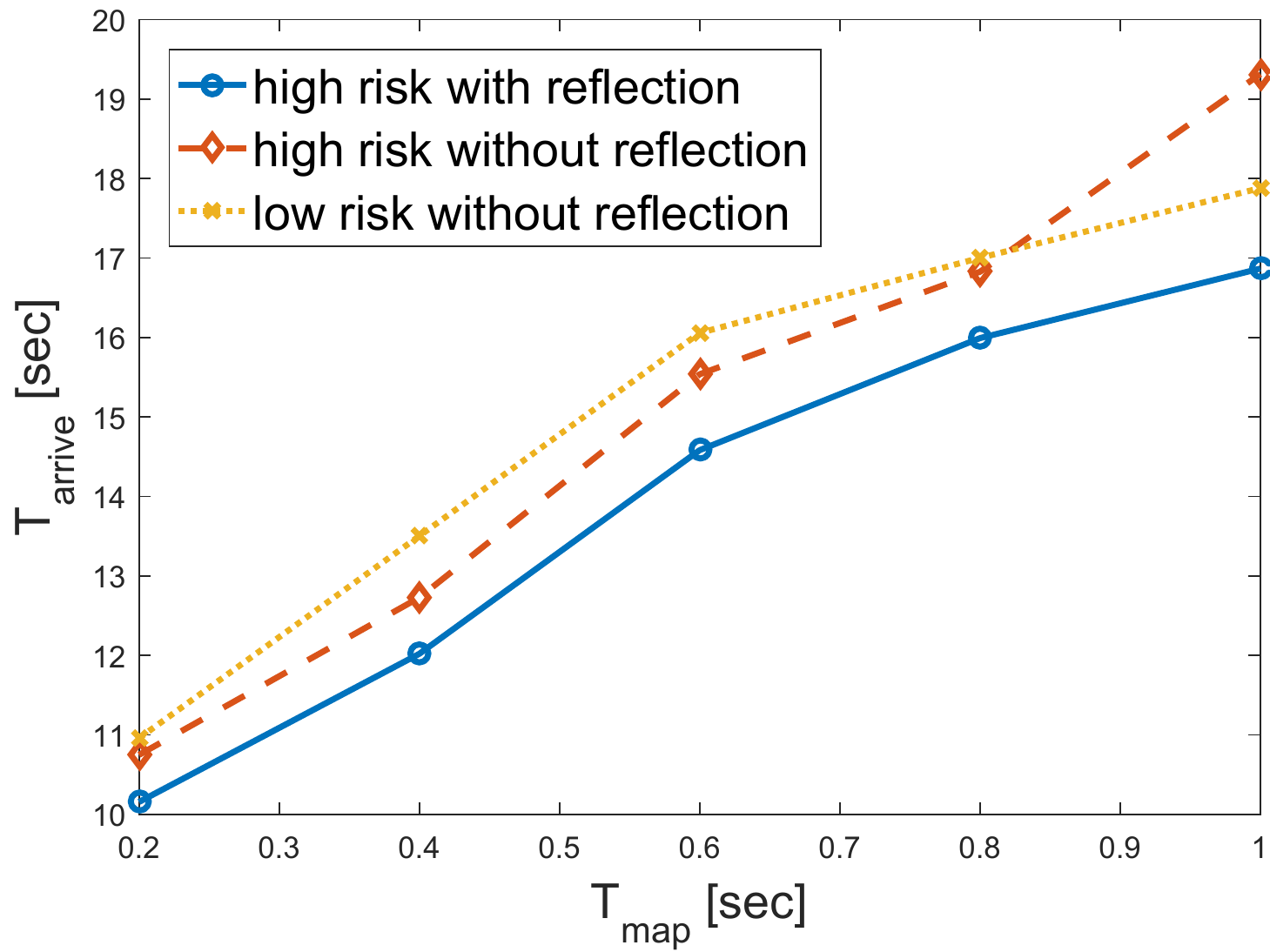}
  \vspace{-3pt}
  \caption{Mean of arrival time}
 \end{subfigure}%
 \begin{subfigure}{.235\textwidth}
  \centering
  \includegraphics[width=0.95\linewidth]{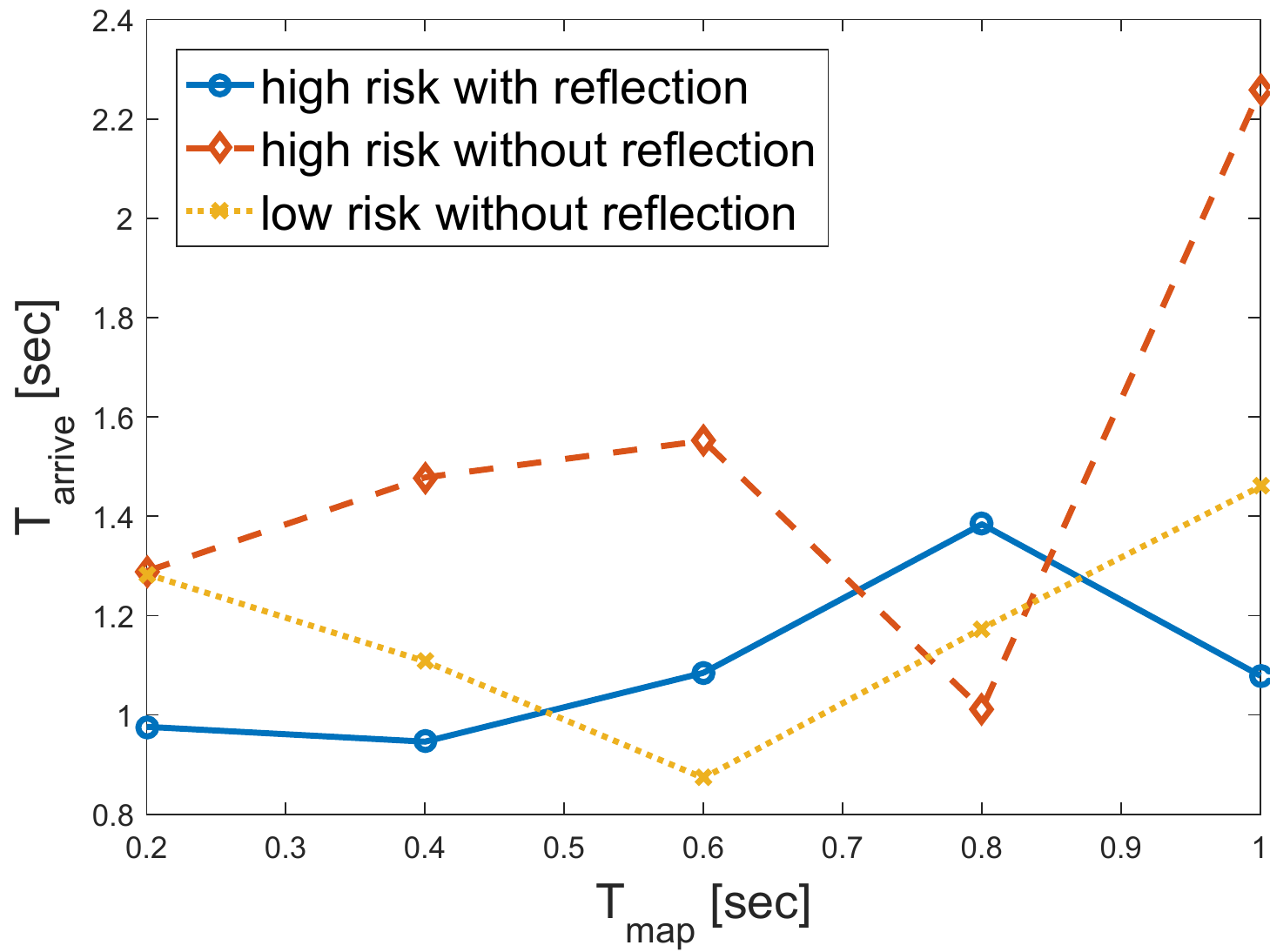}
  \vspace{-3pt}
  \caption{STD of arrival time}
 \end{subfigure}
 \vspace{-1pt}
 \medskip
 \begin{subfigure}{.235\textwidth}
  \centering
  \includegraphics[width=0.95\linewidth]{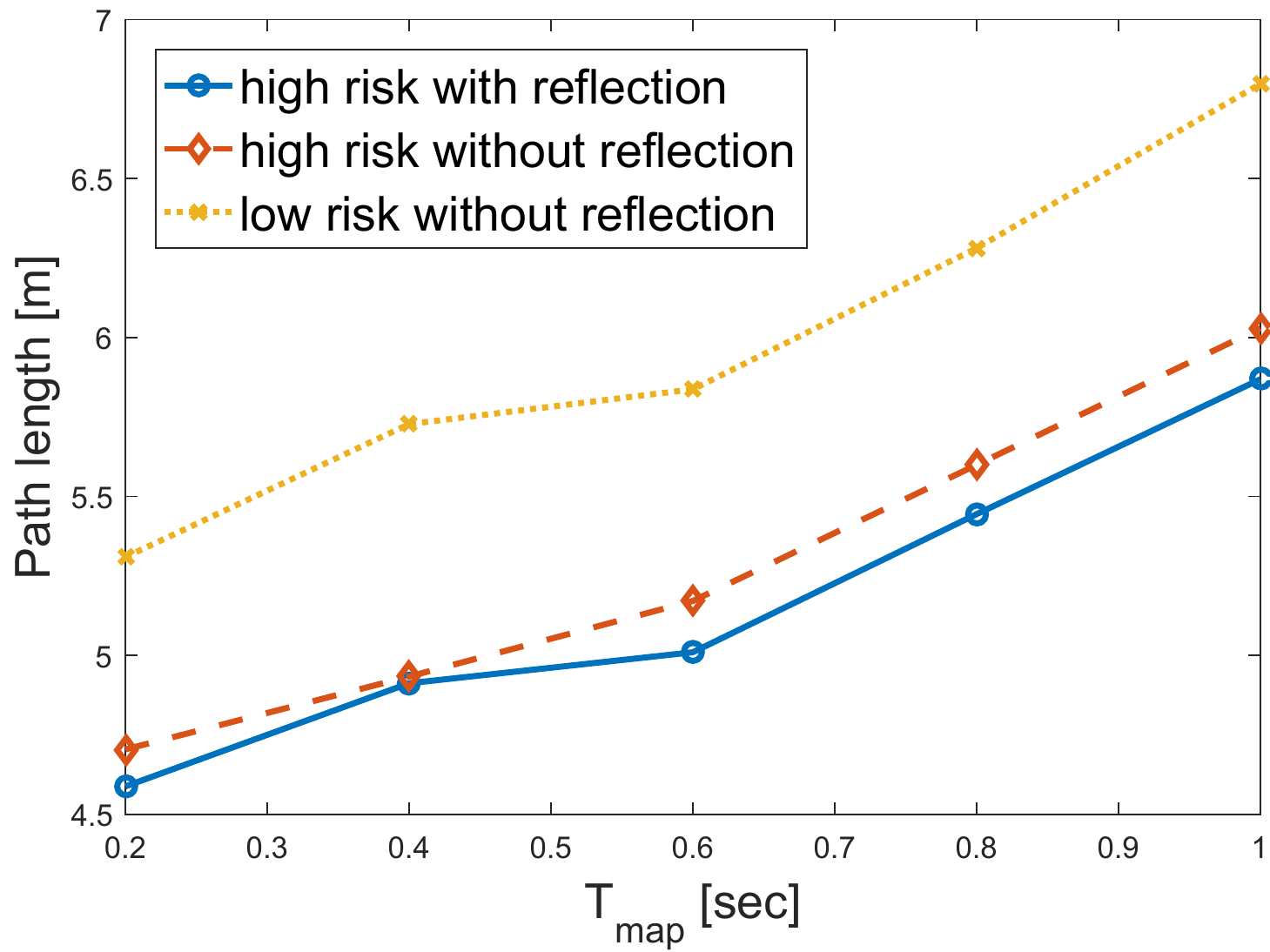}
  \vspace{-3pt}
  \caption{Mean of path length}
 \end{subfigure}%
 \begin{subfigure}{.235\textwidth}
  \centering
  \includegraphics[width=0.95\linewidth]{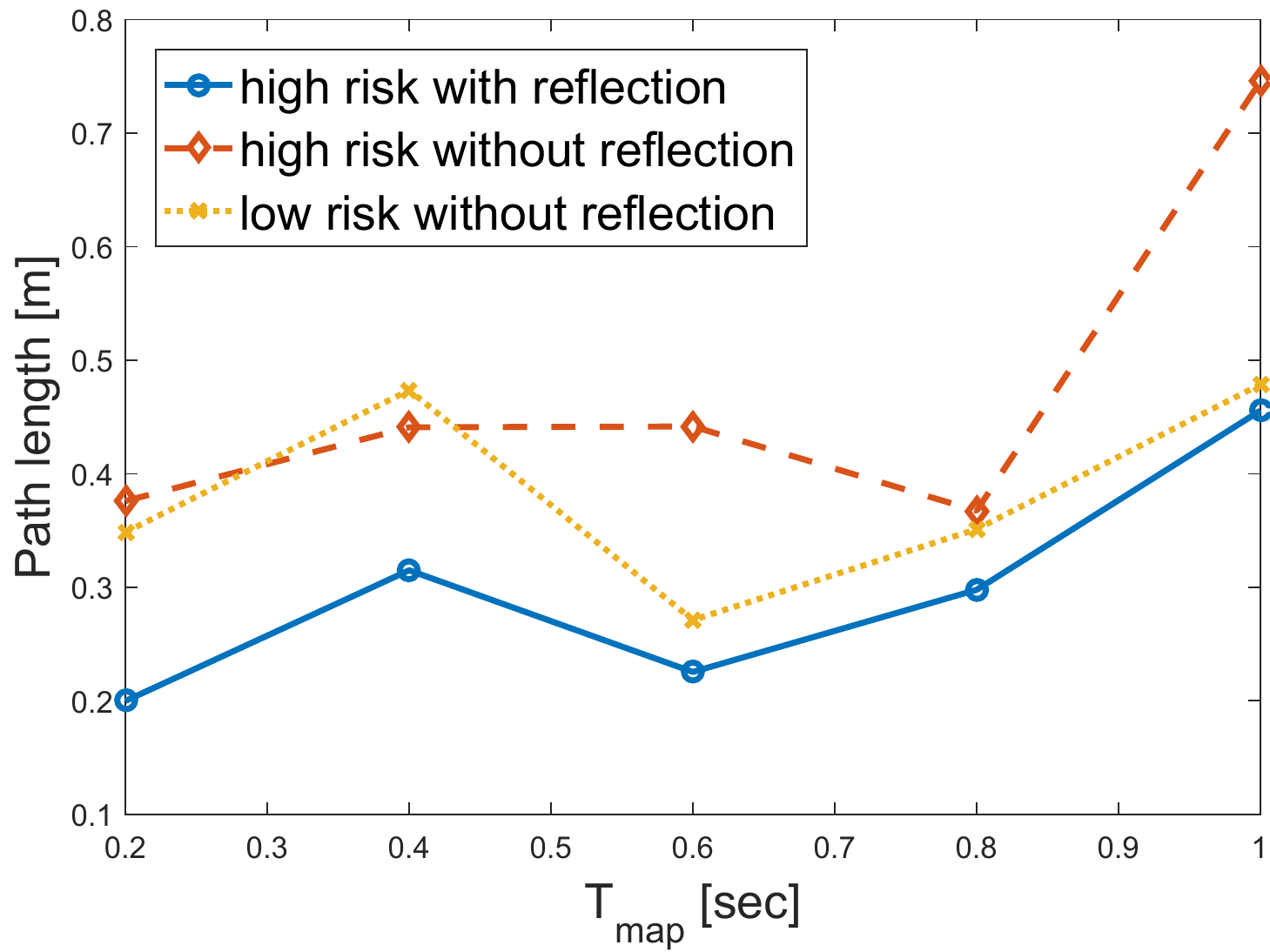}
  \vspace{-3pt}
  \caption{STD of path length}
 \end{subfigure}%
 \vspace{-6pt}
 \caption{Arrival time and path length with different $T_{map}$.}
 \label{fig:parametric study}
 \vspace{-15pt}
\end{figure}

Finally, we investigate the effect of map update time $T_{map}$. We collect $20$ experimental trials for each of the three strategies, and for $T_{map} \in [0.2, 0.4, 0.6, 0.8, 1.0]$\;sec in the same environment, thus giving rise to a total of $300$ experimental paths. 
Results are shown in Fig.~\ref{fig:parametric study}. 
As $T_{map}$ increases, both mean arrival times and mean total path lengths for all three strategies increase. This is because uncertainty in the controller increases as the duration to update the map increases. In all cases, harnessing collisions yields better results; however,  the rates of increase in arrival times and total path lengths do not appear to depend on the employed planning strategy. 
%

\section{Conclusions}
We present a new reactive planning algorithm suitable for collision-resilient robots in unknown environments. The planner allows for potential collisions to be harnessed to improve robot navigation in obstacle-cluttered environments. 
Experiments show that integrating collision harnessing in online planning can decrease both the arrival time and path length of generated trajectories when compared to high-risk trajectories without utilizing collisions ($>7\%$, $>2\%$), and risk-averse safe trajectories that avoid collisions altogether ($>8\%$, $>16\%$). At each planning interval, the pruning technique can save about $3.1 ms$ of computational time.

The work herein contributes foundational algorithmic tools to support the emerging paradigm shift where robots might deliberately choose to collide with the environment, should that help them make progress toward an assigned goal they are tasked to reach. In addition, the pruning technique introduced in this work can apply to other planning algorithms as well, to help reduce the computational time in support of online operation. Future work will focus on analyzing in detail the effect of the various quantities that determine the degree of pruning. We will also focus on experimental evaluation in more complex environments.

	\bibliographystyle{IEEEtran}
	\bibliography{Citation}
	
	

\end{document}